\documentclass[preprint,12pt]{elsarticle}

\usepackage{amssymb}

\usepackage{makecell}
\usepackage{adjustbox}
\usepackage{amsmath}
\usepackage{subcaption}

\journal{}

\begin{document}

\begin{frontmatter}



\title{Customer Profiling, Segmentation, and Sales Prediction using AI in Direct Marketing}


\author[inst1]{Mahmoud SalahEldin Kasem}
\author[inst2]{Mohamed Hamada}
\author[inst3]{Islam Taj-Eddin}

\affiliation[inst1]{organization={Multimedia Department },
            addressline={Assiut University}, 
            city={Assiut},
            postcode={71515}, 
            state={Assiut},
            country={Egypt}}
\affiliation[inst2]{
    organization={Department of Information System, International IT University,},
    city={Almaty},
    postcode={050000}, 
    country={Kazakhstan}
}            
\affiliation[inst3]{organization={Information Technology Department },
            addressline={Assiut University}, 
            city={Assiut},
            postcode={71515}, 
            state={Assiut},
            country={Egypt}}

\begin{abstract}
In an increasingly customer-centric business environment, effective communication between marketing and senior management is crucial for success. With the rise of globalization and increased competition, utilizing new data mining techniques to identify potential customers is essential for direct marketing efforts. This paper proposes a data mining preprocessing method for developing a customer profiling system to improve sales performance, including customer equity estimation and customer action prediction. The RFM-analysis methodology is used to evaluate client capital and a boosting tree for prediction. The study highlights the importance of customer segmentation methods and algorithms to increase the accuracy of the prediction. The main result of this study is the creation of a customer profile and forecast for the sale of goods.

\end{abstract}

\begin{keyword}

 Data mining \sep  SVM \sep  Boosting tree\sep   RFM-analysis methodology\sep Deep learning

\end{keyword}

\end{frontmatter}


\section{Introduction}
\label{sec:Introduction}
In today's business landscape, companies are faced with the challenge of identifying potential customers who are most likely to respond positively to a product or offer, this is where data mining techniques come into play. With the increasing amount of data available, data mining has become an essential tool for direct marketing efforts, allowing companies to create a prediction response model based on past client purchase data. This study aims to present a data mining preprocessing method for developing a customer profiling system that improves the sales performance of an enterprise. The study uses an RFM-analysis methodology to evaluate client capital and a boosting tree for prediction. Furthermore, the study highlights the importance of customer segmentation methods and algorithms in increasing the accuracy of the prediction. The main result of this study is the creation of a customer profile and forecast for the sale of goods, which will assist decision-makers in making strategic marketing decisions. The study is expected to provide valuable insights for companies looking to improve their direct marketing efforts and increase sales performance through data mining-based customer profiling.

The need for a client profiling framework utilizing AI techniques has become increasingly important in today's business landscape. With the exacerbation of competition, the rise in communication costs, and the impact of a lack of buyers, companies are shifting their focus from attracting new customers to retaining existing ones and building their loyalty. The significance of this research topic lies in the fact that long-term relationships with customers are financially beneficial, as they ensure regular purchases, require lower advertising costs per customer and through the recommendations of loyal customers, increase their number. The purpose of this study is to develop a customer profiling system using machine learning methods that will improve the computerized degree of advertising, sales development, and another degree of client base.

To achieve this goal, it is planned to solve the following research tasks:

\begin{itemize} 

\item Data collection; 
\item Study of machine learning methods; 
\item Specify the structure of the client profile, types, and indicators that characterize them; 
\item Analysis and formation of customer data; 
\item Generalize and systematize foreign experience in improving the profile of clients;
\item Conduct an analysis of existing methods for researching the profile of clients, and identify the most effective ones for enterprises;
\item Determine the place and role of the concept of “consumer loyalty” in modern marketing and identify the problems of its use;
\item Clarify the structure and nature of consumer loyalty, types and indicators that characterize them;
\item Generalize and systematize foreign experience in improving;
\item Highlight the factors that determine the choice of a reward system for the formation of programs to increase consumer loyalty;
\item Propose a methodology for developing programs to increase comprehensive consumer loyalty for manufacturers of goods and services and formulate practical recommendations for their formation.

\end{itemize}

Deep learning is a subfield of machine learning that has seen widespread applications in various industries. In computer vision, deep learning algorithms have been utilized for object detection, image classification, and video analysis. In the field of Natural Language Processing (NLP), deep learning models have been applied to tasks such as text classification, sentiment analysis, machine translation, speech recognition\cite{chorowski2015attention}, table detection and recognition\cite{abdallah2022tncr,prasad2020cascadetabnet,kasem2022deep}. Healthcare is another industry where deep learning has found several applications, including diagnosis, treatment planning, drug discovery\cite{fakoor2013using}, and medical imaging analysis\cite{nie2015disease,abdallah2020automated,yu2014deep}. In robotics, deep learning is used for autonomous navigation, object recognition\cite{logothetis1996visual,nurseitov2022application}, and robotic control. handwritten recognition for various languages\cite{mahmoud2014khatt,nurseitov2021handwritten,toiganbayeva2022kohtd,2020_hkr,DaniyaAbdallah2020}. Intrusion Detection in IoT \cite{mahmoud2022ae,xu2021improving} The finance industry has also seen applications of deep learning in areas such as fraud detection, algorithmic trading, and risk management. Additionally, deep learning is finding use cases in gaming, such as game playing and decision-making, as well as in marketing, with applications in customer segmentation, personalized recommendations, and sentiment analysis. The transportation industry is another area where deep learning is making an impact, with applications in autonomous vehicles, traffic prediction, and route optimization. Finally, deep learning has potential applications in the energy industry for predictive maintenance, energy consumption prediction\cite{waschneck2018optimization,hamada2021neural}, and equipment malfunction detection. These are just some of the ways that deep learning is being applied across different fields and the potential for further growth and development is immense.

The objects of study are enterprises and organizations, their marketing activities in the context of the formation and implementation of client policy, as well as consumers of goods and services. The subject of this study is the entirety of economic and organizational relationships that occur in the process of firms implementing relationship marketing, as manifested in the creation and implementation of programs to build consumer loyalty.

The study's theoretical and methodological foundation was the essential research of internal and international scientists on issues of market economy, management, marketing, and consumer and brand loyalty management\cite{zopounidis2002new}. The methods of marketing, economic and statistical analysis, quantitative and qualitative study, as well as the principles of consistency and development, were used in work. The author also relied on substantiating the main provisions of the dissertation on expert methods for obtaining information. As described by the authors of the article, H. Muller and U. Hamm, the first step is to start with segmentation, marketing, and customer data. Then, the data can be adjusted in the right direction for analysis and profiling\cite{muller2014stability}.

The scientific novelty of the work lies in the development of scientific and methodological provisions and recommendations aimed at the formation and implementation of a client profiling framework utilizing AI techniques, as well as the identification of the most effective methods for researching the profile of clients and increasing consumer loyalty in Kazakhstani enterprises. This study will provide valuable insights for companies looking to improve their relationship marketing efforts and increase sales performance through data-driven customer profiling.

The research aims to analyze and review various projects, works, and scientific literature on the topic of customer segmentation in online business ventures. The overall description of the research is that in online business, clients use various platforms provided by organizations with various needs, shopping patterns, and profiles. To understand this wide range of needs, shopping patterns, behavior, conduct, and requests of clients, we use various divisions according to the business model of organizations. Customer segmentation is defined as the division of customers into various individual groups that share similarities in various ways relevant to marketing, such as orientation, interests, age, shopping patterns, and different ways of managing money.

Organizations that want to implement customer segmentation are under the idea that different customers have different needs and requirements, which is why organizations perform data mining procedures and develop a specific marketing strategy to implement in their business model. The simple truth is that most organizations have data that can be used to target these individuals and to understand the critical drivers of segmentation. Customer segmentation is when a customer is divided between various groups based on business needs.

Today, customers have become the fuel that drives a business. The loss of customers affects sales, which is why it is expensive to acquire new customers. At the same time, it is more important to retain old customers. Therefore, organizations need to focus on reducing customer churn and to do so, they constantly offer coupons and offers to customers. It is good to know that AI will help with customer segmentation. One of the main uses of unsupervised learning methods of AI is customer segmentation. With the help of the “clustering” procedures of unsupervised learning, we can identify the different segments of customers. Where individual segments have some individual similarities, which allows organizations to target the potential customer base according to the business model and requirements for efficiency.

The research will also cover the need for customer segmentation, the importance of understanding customer behavior, and the use of AI in customer segmentation. The study will provide valuable insights for organizations looking to improve their customer retention and benefit upgrades through data-driven customer segmentation.

\section{Related Work}
\label{sec:Related Work}

In the field of customer segmentation, researchers have been experimenting with different algorithms to perform segmentation on customer data. Most of these studies have focused on analyzing customer buying history and purchasing behavior to identify segments.

According to T Jiang and A, Tuzhilin \cite{jiang2008improving}, it is crucial to implement both customer segmentation and buyer targeting in order to enhance marketing performance. These two tasks are integrated into a step-by-step approach, however, the challenge of unified optimization arises. To address this issue, the authors proposed the K-Classifiers Segmentation algorithm. This method prioritizes allocating more resources to those customers who generate the most returns for the company. A significant number of researchers have discussed various techniques for segmenting customers in their studies. also, the authors propose a direct clustering method for grouping customers. Rather than relying on computed statistics, this approach utilizes transactional data from multiple customers. The authors also acknowledge that finding an optimal segmentation solution is computationally difficult, known as NP-hard. Therefore, Tuzhilin presents alternative sub-optimal clustering methods. The study then experimentally evaluates the customer segments obtained through direct grouping and finds them to be superior to statistical methods.

KR Kashwan \cite{kashwan2013customer} proposed a K-means algorithm and a statistical tool to propose a model that elaborates on a continuous analysis and online framework for an e-commerce organization to predict sales. They involved a clustering strategy for determining market segmentation because a developed computing-based system is intelligent enough to address results to managers for a quick and fast decision-making cycle.

PQ Brito \cite{brito2015customer} emphasized that advertising and manufacturing approaches are highly important for customized industries because buying a large variety of products makes it difficult to find specific patterns of customer preferences. As a result, they proposed two different data mining methods, clustering and sub-cluster discovery, for customer segmentation to better understand customer preferences.

X He and C Li \cite{he2016research} propose a three-dimensional strategy for enhancing customer lifetime value (CLV), customer satisfaction, and customer behavior. The study concludes that consumers have varying needs, and segmentation helps to identify their demands and expectations, which in turn leads to providing better service.

A Sheshasaayee \cite{sheshasaayee2017efficiency} developed a new integrated approach to segmentation by combining the RFM (Recency, Frequency, Monetary) and LTV (Life Time Value) methods. They employed a two-phase approach, starting with a statistical method in the first phase, and then proceeding to cluster in the second phase. The objective is to apply K-means clustering following the two-phase model and then utilize a neural network to improve the segmentation.

MT Ballestar \cite{ballestar2018customer} proposed the role of customers in the use of their cashback and determined the business activity and behavior of customers on the site of a social network. They proposed a model that applied social network analysis to marketing such as loyalty, communication, customer development, and customer engagement to show the dependence of customers’ positions within an organization.

W Qadadeh \cite{qadadeh2018customers} proposed the evaluation of data analysis algorithms like K-means for clustering and Self-Organized Maps for the nature of clustering with visualization. They recommend that involving various procedures for segmentation with experts will further develop organizations like insurance and study segment elements and behavior of a customer in any customer relationship management dataset.

AJ Christy \cite{christy2021rfm} emphasized that a good understanding of the customer’s needs and identification of potential customers for the organization are satisfied by the segmentation process. They performed segmentation using RFM analysis and extended it to other algorithms like K-means, and RM K-means through minor adjustments in K-means clustering.

\begin{table}[h!]
\begin{center}
\caption{\footnotesize{ Related work methods, advantages and disadvantages }}
\begin{adjustbox}{width=1\textwidth}
\begin{tabular}{|c|c|c|c|}
\hline

\textbf{Paper} & \textbf{Proposed Method} & \textbf{Advantages} & \textbf{Disadvantages} \\
\hline

KR Kashwan \cite{kashwan2013customer} & \makecell{K-means algorithm an-\\d a statistical tool } & \makecell{A continuous analysis and \\online system for e-comm-\\erce organization to pred-\\ict sales}  & \makecell{ Limited to the use of clustering \\strategy for determining of market \\segmentation} \\ \hline

PQ Brito \cite{brito2015customer} & \makecell{Two data mining met-\\hods (clustering and \\sub-cluster discovery) }& \makecell{Better understanding of \\customer preferences } & \makecell{ Limited to redefined \\industries}\\ \hline

MT Ballestar \cite{ballestar2018customer} & \makecell{ Utilization of cashback \\ and client behavior on \\social network sites} & \makecell{Shows the reliance on\\ the position of clients \\inside an organization }& \makecell{Limited to the use of\\ social network writing\\ to promoting like dedication, \\person-to-person communication,\\ development of client,\\ and commitment of client } \\ \hline

W Qadadeh \cite{qadadeh2018customers} & \makecell{K-means for clustering \\and Self-Organized Maps\\ for quality of clustering\\ with representation} & \makecell{ Involves various proced-\\ures for division with ex-\\pert to further develop\\ organizations } & \makecell{ Limited to the use of \\multiple procedures for\\ segmentation with expert} \\ \hline

AJ Christy \cite{christy2021rfm} & \makecell{RFM analysis and exte-\\nded to other algorithms \\like K-means, and RM\\ K-means }&\makecell{ Good understanding of \\the need of client and \\identification of potential \\clients for organization } & \makecell{ Limited to the use of\\ RFM analysis and extended \\it to other algorithms like \\K-means, and RM K-means \\through minor adjustment in \\K-means clustering} \\ \hline

T Jiang  \cite{jiang2008improving} & \makecell{Direct clustering based \\on transactional data} & \makecell{Identifies customer seg-\\ments based on actual \\customer behavior} &  \makecell{Finding an optimal\\ segmentation solution is \\computationally difficult} \\ \hline

X He  \cite{he2016research}   & \makecell{Three-dimensional appr-\\oach for enhancing CLV, \\customer satisfaction, \\and customer behavior} & \makecell{Considers multiple dim-\\ensions of customer be-\\havior, leading to more \\accurate segmentation} &  \makecell{Complexity and High \\computational cost} \\ \hline

A Sheshasaayee \cite{sheshasaayee2017efficiency}    & \makecell{Integrated approach com-\\bining RFM and LTV me-\\thods with two-phase app-\\roach (statistical and clust-\\ering) and neural network} & \makecell{Integrates different met-\\hods to improve segment-\\ation} &  \makecell{Computationally intensive} \\ \hline

\end{tabular}
\end{adjustbox}
\end{center}
\end{table}

\section{Problem Statement}

The problem of customer segmentation can be based on various factors such as marketing, sales, support, product, and leadership. Experts in large or small organizations involved in the data analysis process adjust the working group and set the expectations that it will continue to do so in many stages. Some issues that can be resolved through customer segmentation are given below.

\begin{itemize} 
\item Marketing: We can solve the problem by understanding our customer base to effectively reach them. We may not be able to observe the business's email lists using the task to be done, but we can observe ones for B2C subscription organizations with high website traffic volume.
\item  Sales: Many issues faced by sales representatives can be resolved by this process. We can route prospects to our self-service stream or the most appropriate group within sales, such as startups, Small Market businesses, and Multi-Model businesses, based on clear customer segments.
\item Support: Issues are categorized based on their tool and field. After categorization, it can be used to route support inquiries to the appropriate channels, such as AnswerBot, Alexa, Google Assistant, our help center, or a support representative, to improve customer and business outcomes further.
\item  Product: This process can also resolve issues with product quality. Experts should know which product requests and feedback make the biggest impact on which customer and focus accordingly, instead of by volume alone.
\item  Leadership: It manages the mission run by e-commerce organizations to deliver their service and make a lead. For this, they create a common language for the product and design and go to markets to describe the customers.

\end{itemize}

From this work, we proposed a customer segmentation strategy based on various categories. Different clustering methods like k-means, RM K-means, and Self-Organized Maps were used for segmentation. In this paper, we proposed a business model for e-commerce organizations based on segmentation according to various categories [10] and RFM positioning to retain and acquire customers in e-commerce. As we know, observing new customers is important, but retaining old customers is even more important.

\section{MODEL, TOOLS, ENVIRONMENT, TECHNOLOGY}
\label{sec:MODEL}

\subsection{The Customer Segmentation Approach}
Client division is a commonly used marketing technique where a business separates its customer base into smaller groups that can be targeted with specific content. This is done by analyzing customer behavior data, which gives the company a deeper understanding of the types of customers in its system. The benefit of this technique is that it allows for more effective marketing strategies. It can be challenging to implement in online retail environments, where data is vast and complex. One algorithm used for this purpose is Vector Quantization, which automatically assigns existing or new customers to different groups. It was developed by Linde, Buzo, and Gray and can be applied to any probability source definition or long data sequence. It may not always achieve optimal results, but it often ensures local optimality \cite{pranata2015segmenting}.

Client segmentation is a powerful marketing strategy that is widely used by businesses to understand better and target their customer base. It involves dividing the customer base into smaller groups based on characteristics such as demographics, behavior, and purchasing history. These segments are then targeted with specific marketing messages and campaigns that are tailored to their unique needs and preferences.
One of the key advantages of client segmentation is that it allows businesses to understand better the different types of customers that make up their customer base. This, in turn, leads to more effective marketing strategies that are better able to convert leads into customers. Additionally, segmentation allows businesses to identify and target their most valuable customers, which can help to improve customer retention and increase revenue.
Implementing client segmentation can be challenging, particularly in online retail environments with vast and complex data. One algorithm used for this purpose is Vector Quantization, which automatically assigns existing or new customers to different groups. It was developed by Linde, Buzo, and Gray and can be applied to any probability source definition or a long data sequence. This algorithm is efficient and can automatically group customers based on their behavior data.
Although Vector Quantization is an efficient algorithm and a powerful tool for client segmentation, it may not always achieve optimal results, but it often ensures local optimality. In most cases, the algorithm is able to accurately group customers and provide valuable insights for businesses to target their marketing efforts.

A mapping, also known as a block quantizer or vector quantizer, is a tool that can be used to divide data into smaller groups. The mapping is N-level k-dimensional and can be implemented in software code. It can take a variety of client RFM values as input vectors, and a non-negative real distortion measure is used to represent the difference between the original vectors and the reproduced vectors. This type of mapping has been widely studied in the literature, and many similar techniques have been presented \cite{pranata2015segmenting}. The error distortion measure, which is widely used in the development of mathematical applications, is chosen for its computational efficiency in the formula.
\begin{equation}
\begin{split}
 d(x,x^n)  = \sum_{i=0}^{k-1} | x_i - x_i^n | 
\end{split}
\end{equation}

Where x is the input vector, $x^n$ = q(x) is the reproduction vector, n describes the number of times the division, and i is the reproduction vector.

An N-level quantizer is considered to be ideal or globally ideal if it minimizes the average distortion, or at least, is ideal if for any remaining quantizers four having N generation vectors D(q*)<D(q) \cite{christy2021rfm}. If D(q) is only a close least, a quantizer is said to be locally optimal, meaning that small changes in q cause distortion to increase. The goal of quantizer design is to obtain an optimal quantizer if possible or a locally optimal and preferably "good" quantizer if that is not possible. Several such algorithms have been proposed in the literature for the computer-aided design of locally optimal quantizers.

\subsection{Machine Learning}
Recently, interest in machine learning (ML) has grown as processing power and accumulated data have increased significantly. Artificial intelligence (AI) can be defined as "computational methods that make use of past experience to improve performance or make precise predictions." Experience, in this case, refers to data about the past, which is often electronic information, the size and quality of which significantly impact the outcome of the predictions made by the algorithms. Common AI tasks include classification, regression, ranking, clustering, dimensionality reduction, or complex learning. Classification is a problem of finding the right label for inputs. These problems can be, for example, image labels, text classification, or identifying the appropriate customer segment for a customer. Regression is a problem where a value is not fixed for an input. For example, future stock value or length of the customer relationship. In Ranking, the problem is to arrange items according to certain criteria, for example, web search. Clustering aims to segment the data into homogenous groups that are not yet known. For example, a company might wish to discover new customer segments or in social networks to find communities. Dimensionality reduction or complex learning aims to reduce data representation to lower layered representation. The subject of this review is whether a customer will be stirred, which is a common decision problem between 1 and 0. Therefore, the strategies introduced in this section are used in decision problems. AI techniques can be divided into supervised learning and unsupervised learning, where the main difference is that with supervised learning, the data is labeled. In unsupervised learning, it is not. A common use case for unsupervised learning is clustering or dimension reduction and, for example, email spam filter for supervised learning.

\subsubsection{Data preprocessing and model optimization}
Data preprocessing is an essential step in creating an AI model. It plays a crucial role in the model's performance and its interpretability. Data preprocessing includes cleaning, standardization, transformation, feature extraction or selection, and more. The preprocessing can be divided into two main categories: value transformation (cleaning, standardization, transformation, handling missing values, etc.) and value representation (variable selection and evaluation).

\subsubsection{Data cleaning and transformation}
Data cleaning is examining the quality of data and ensuring its integrity. This process involves two primary approaches: filtering and wrapping. Filtering involves the removal of data based on predefined rules, such as removing outliers, incorrect spellings, duplicates, or impossible data, such as a 120-year-old client. On the other hand, Wrapping focuses on improving data quality by identifying and removing mislabeled data.
Feature engineering or data transformation is a technique used to discover missing information about the relationships among features and construct new features from the existing features. This process can lead to more accurate and concise classifiers and improved interpretability. These new features may include combinations of current and future values, such as the sum of two previous values.

\subsubsection{Missing Data}
Data preprocessing often involves dealing with missing values in the dataset. One approach to handling missing data is to delete the instances that contain missing values, which can lead to data imbalance. An alternative approach is to impute the missing values with an estimated value. This can be done by using similar instances, calculating the mean values, or using statistical or machine learning techniques.

\subsubsection{Sampling}
Data preprocessing is a crucial step in creating an AI model. It affects the performance and interpretability of the model. It includes data cleaning, standardization, transformation, feature extraction, and selection. Data cleaning involves examining the quality of the data and removing any anomalies or inaccuracies. Transformation or feature engineering is a technique used to find missing data and build new features from existing ones that would result in more accurate and concise classifiers and increased interpretability. Missing values can be handled by imputing estimated values obtained from similar cases or using statistical or AI techniques. However, it is important to note that class imbalance, a common issue in ML, can lead to problems such as improper evaluation metrics, lack of data, and improper inductive bias. To address this, oversampling and undersampling techniques can be used to adjust the distribution of the training set. Still, they also have their own drawbacks, such as loss of data and increased risk of overfitting.

\subsubsection{Feature and Variable Selection}
Feature and variable selection are techniques used to identify and extract the most relevant data from many factors. With the increasing amount of data and factors available due to advancements in data collection, it is crucial only to include the most important and useful factors in the model being built. The main goals of selection are to achieve better predictive performance, make faster and more efficient predictions, and gain a more accurate understanding of the predictive process. Including unnecessary factors in the model can lead to complexity or overfitting, while missing important factors can result in diminished predictive performance.
There are several classes of feature selection methods, including filter, wrapper, and embedded methods. Filter methods use selected feature importance models, such as variance, to determine which features to include. Wrapper methods use algorithms to iterate through possible feature subsets and maximize classification performance, while embedded methods aim to reduce computational time by incorporating feature selection into the training process. Advanced methods, such as genetic algorithms and particle swarm optimization, can also be used. However, these methods can be computationally expensive and may be subject to NP-hard problems. It's important to use appropriate feature selection methods based on the dataset, model, and computational resources.

\subsection{Model for Customer Segmentation}
Several models are commonly used for customer segmentation, including classification techniques and various analytical methods tailored to the specific needs of different business models. The main models used for customer segmentation include:
\begin{itemize}
    \item Demographic Segmentation;
    \item Recency, Frequency, and Monetary (RFM) Segmentation;
    \item Customer Status and Behavioral Segmentation.
\end{itemize}

Segmentation based on gender is one of the simplest yet most effective ways for organizations to categorize their customer base. This type of segmentation is particularly useful for creating targeted content or promotions for gender-based events or programs, such as Mother's Day, Father's Day, or Women's Day.
RFM Segmentation is commonly used in the direct mail industry and is widely employed for ranking customers based on their purchasing history. This approach identifies customers based on recency (the number of days between two purchases), frequency (the total number of purchases made by a customer in a specific period), and monetary value (the total amount spent by a customer in a specific period)\cite{brito2015customer}.

Client status and behavior analysis is when organizations examine their data to categorize their clients into active and lapsed. Active and lapsed status refers to the last time a client made a purchase\cite{tsao2019product}. Behavioral analysis involves analyzing the past behavior of clients, such as shopping habits, brand preferences, and purchase patterns, to make predictions about their future actions. This process is carried out by data analysts who work with the data set from the e-commerce organization, load the data, perform data analysis, and segment the clients into categories. The information is then presented in easy-to-understand dashboards for non-technical individuals. Finally, this information is used to develop strategies for retaining and acquiring clients.

Brain networks are a component of Artificial Intelligence that employs principles and behavior similar to that of neurons in living organisms for signal processing \cite{chen2005mining}. The central aspect of this network, which accounts for its broad possibilities and significant potential, is the parallel processing of data by all nodes, significantly enhancing the speed of data processing. Additionally, with a high number of interneuron connections, the network possesses robustness against errors that may occur in individual lines. Currently, brain networks are applied to solving various problems, one of which is the problem of prediction\cite{witten2005practical}. In this case, the radial basis function (RBF) network was chosen as the architecture of the brain network, with a multi-layered time series as input and the prediction outcome as the time series value at the desired time.

To improve the prediction quality, it is crucial to perform preprocessor data handling, as brain networks typically do not perform well with values from a broad range of input data. To eliminate this issue, the data should be scaled to the range [0... +1] or [-1... +1]. The equation used to scale the input data is as follows (2,3,4):

\begin{equation}
\begin{split}
 X_s = S_c . X_u + Of
\end{split}
\end{equation}

\begin{equation}
\begin{split}
Of = \frac{T_{max} - T_{min}}{R_{max} - R_{min}}
\end{split}
\end{equation}

\begin{equation}
\begin{split}
 Of = T_{min} - S_c . R_{min}
\end{split}
\end{equation}

Where Xs, Xu respectively, the scaled and original input data;
Tmin=0, Tmax=1 - the maximum and minimum of the objective function;
Rmax, Rmin - the maximum and minimum inputs.

\begin{figure}[ht!]
      \centering
      \includegraphics[width=\textwidth,height=.60\textwidth]{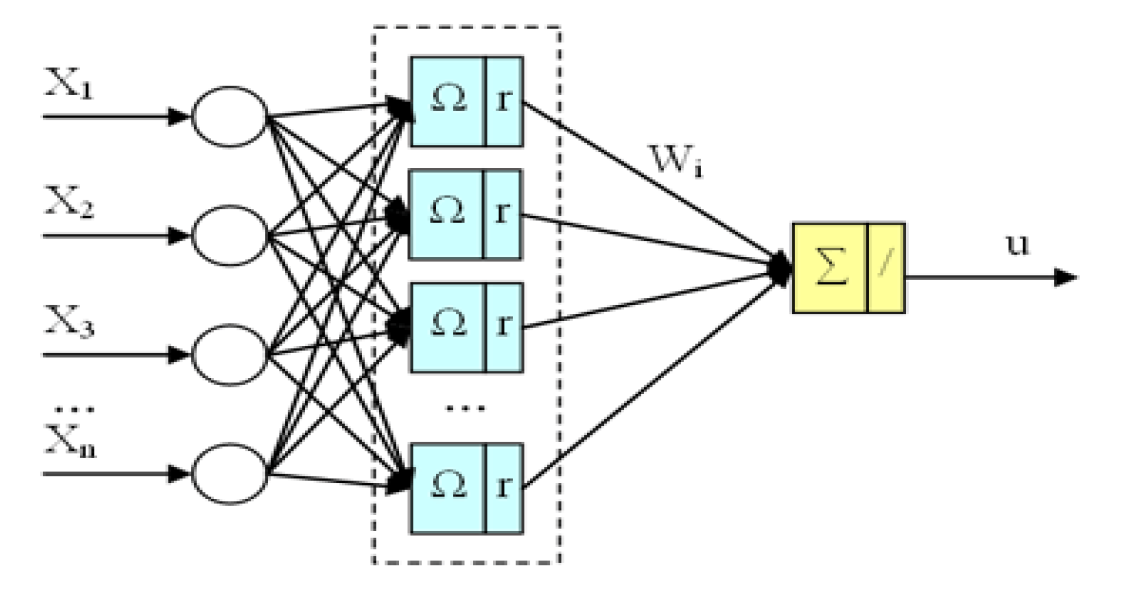}
      \caption{Radial basis neural network}
      \label{fig:Radial-basis}
\end{figure}

A radial basis neural network is a network with one hidden layer in figure \ref{fig:Radial-basis}

In the work context, the hidden layer employs Radial Basis Functions (RBFs) to transform the input vector X. various radial basis functions can be utilized. However, the Gaussian function is the most commonly used and will be utilized in this work. The Gaussian form for the kth neuron is as follows \cite{shaw2001knowledge}:

\begin{equation}
\begin{split}
 \phi_k(x) = exp(\frac{-r_k^2}{a_k^2})
\end{split}
\end{equation}
where X is the input vector, $ r_k$ is the radius.

\begin{equation}
\begin{split}
  r_k = | X - C_k |
\end{split}
\end{equation}

$C_k$ is the center vector of the RBF, and a is the function's parameter, called the width. The output layer of the network is a linear adder, and the output of the network $C_k$ is described by the expression:

\begin{equation}
\begin{split}
  u = \sum_{k=0}^{N} w_k  \phi_k(X) 
\end{split}
\end{equation}

where wk is the weight connecting the output neuron with the kth neuron of the hidden layer.

To understand the behavior of a radial basis function network, tracking the progression of the input vector X is crucial. When values are assigned to the components of the input vector, each neuron of the input layer produces a value based on how close the input vector is to the weight vector of each neuron. Consequently, neurons with weight vectors that differ significantly from the input vector X will have outputs close to 0, and their impact on the results of subsequent neurons in the output layer will be negligible. Conversely, an input neuron whose weights are close to the X vector will produce a value close to one. Segmentation is performed on an unstructured set of customer data intended for the purpose of marketing. This section discusses market segmentation and customer segmentation and mentions the available data mining techniques to support these processes. Market segmentation is a well-known marketing strategy, and its benefits are highlighted in various marketing research textbooks\cite{shaw2001knowledge}.

\subsection{Market segmentation}
Market division, first defined in 1956, is a method used by organizations to categorize customers based on similar characteristics, such as geographic location, demographics, product usage, and purchasing behavior. The goal is to increase customer satisfaction and maximize efficiency by tailoring marketing efforts to specific segments. One common tool used in market division is clustering, which groups elements with similar values into segments\cite{liu2012unified,kim2006customer,weinstein2013handbook}.

While early market division studies only considered one set of factors, modern market division models take into account multiple sets of factors simultaneously, called cooperative market division. There are various market division methods, including k-means clustering, hierarchical clustering, association rule mining, decision trees, and neural networks. The objective is to identify and describe customer groups and reach profitable customer segments\cite{hosseini2015new}.
The stages of market division research include Literature Review, Solution Architecture, Testing, Verification, and Evaluation of Results. Market division is an ongoing area of study, and there is always room for improvement. The ultimate goal of using a market division system is to improve the position of the organization and better serve the needs of customers\cite{yankelovich2006rediscovering}.

\subsection{Customer segmentation}
Market and customer segmentation are often used interchangeably in the literature, with market segmentation generally being viewed as a high-level strategy and customer segmentation providing a more granular view. A combination of customer segmentation and targeting for campaign strategies can be achieved through the use of the Recency, Frequency, and Monetary (RFM) model\cite{brito2015customer}. The RFM model considers the most recent purchase amount (P), the total number of purchases made during a given period of time (F), and the monetary value spent during that time period (M). It can be used in conjunction with the Customer Lifetime Value (LTV) model, which evaluates the contributions of segmented customers by calculating their current value and predicting their potential value.

One approach to improve the customer division and targeting process is through the use of genetic algorithms, as proposed by Chang\cite{qadadeh2018customers}, who suggests that the LTV model be used as a fitness function in the genetic algorithm to identify more suitable customers for each campaign. Another approach, proposed by Kim, Jung, Su, and Hwang\cite{kim2006customer}, is to perform customer segmentation using LTV components such as current value, expected value, and customer loyalty.

In traditional markets, customer segmentation is a critical technique used in marketing research. There are numerous mathematical methods for identifying customer segments, including statistical techniques, neural networks, genetic algorithms, and k-means fuzzy clustering, as explored by various researchers.

To conclude, a brief overview of the segmentation process is provided. The customer population can be divided into segments based on different criteria or attributes. For example, a population could be segmented based on geographic location, resulting in four segments of varying sizes. However, the segments would have different attributes that could be further exploited through a process called customer profiling.

\subsection{Client profiling}
Client profiling involves analyzing a client's characteristics such as age, orientation, income, and lifestyle in order to understand the traits of a particular group and describe what they are like. By utilizing client segmentation and profiling techniques, marketers can determine the appropriate marketing strategies for each segment. This approach helps to establish and maintain a strong relationship with existing customers, improving customer retention and ultimately contributing to business growth and revenue generation. This process is known as Customer Relationship Management (CRM) \cite{romdhane2010efficient}. There is no one specific method for conducting client segmentation and profiling, as each database utilizes its own approach. Typically, there are two types of profiling: segment profiling and lead profiling \cite{chan2008intelligent}.

Client segment profiling is a common marketing approach to understanding the attributes of a particular group of customers. It takes into consideration various factors such as demographics, lifestyle, and purchasing behavior to tailor marketing strategies and enhance customer relationships. This practice falls under the umbrella of customer relationship management (CRM) and is crucial for improving customer acquisition and revenue generation in the early stages of a digital project. The segment profile of the customer is considered more relevant than the individual social profile as it determines the target market for advertising and provides insight into the content direction.
Additionally, the decision-making process of consumers in regard to purchasing goods and services is known as buyer behavior. While Mowen and Minor present a different definition, behavior profiling is based on consumer attitudes, usage patterns, and reactions to a product.
Advertisers believe that social factors are the best starting points for constructing consumer behavior profiling, including:
\begin{itemize}

    \item Timing: Customers are profiled based on their purchase decision-making process, including the time they choose to make a purchase or use the product. Companies may adopt different marketing strategies based on key timing events, such as before the New Year or National Holidays.
    \item Benefits: Benefit profiling is a process that segments customers based on the various benefits they may be seeking in a product.
    \item Customer status: By profiling non-customers, former customers, potential customers, new customers, and regular customers of the product, the company can tailor and customize its marketing efforts for each group.
    \item Usage rate: Usage rate profiling segments customers based on the amount they use the product, dividing them into groups of non-users, light users, medium users, and heavy users.
    \item Purchaser Readiness Stage: The purchaser readiness stage refers to the customer's level of awareness and interest in the product.
    \item Loyalty status: Customers can also be profiled based on their level of loyalty. Hard-core loyal customers consistently purchase the same product, split loyal customers are loyal to multiple brands and purchase them randomly, and shift loyal customers switch from one brand to another, staying with one brand for a period of time before switching to another.
    \item Attitude: Customers can be divided based on their attitude towards the product, such as enthusiastic, positive, neutral, negative, or hostile. By considering consumer attitudes towards a brand or product, a company gains a wide range of insights about the market and its customers.

\end{itemize}

\section{EXPERIMENTAL}
\subsection{Architecture}
\begin{figure}[ht!]
      \centering
      \includegraphics[width=\textwidth,height=.60\textwidth]{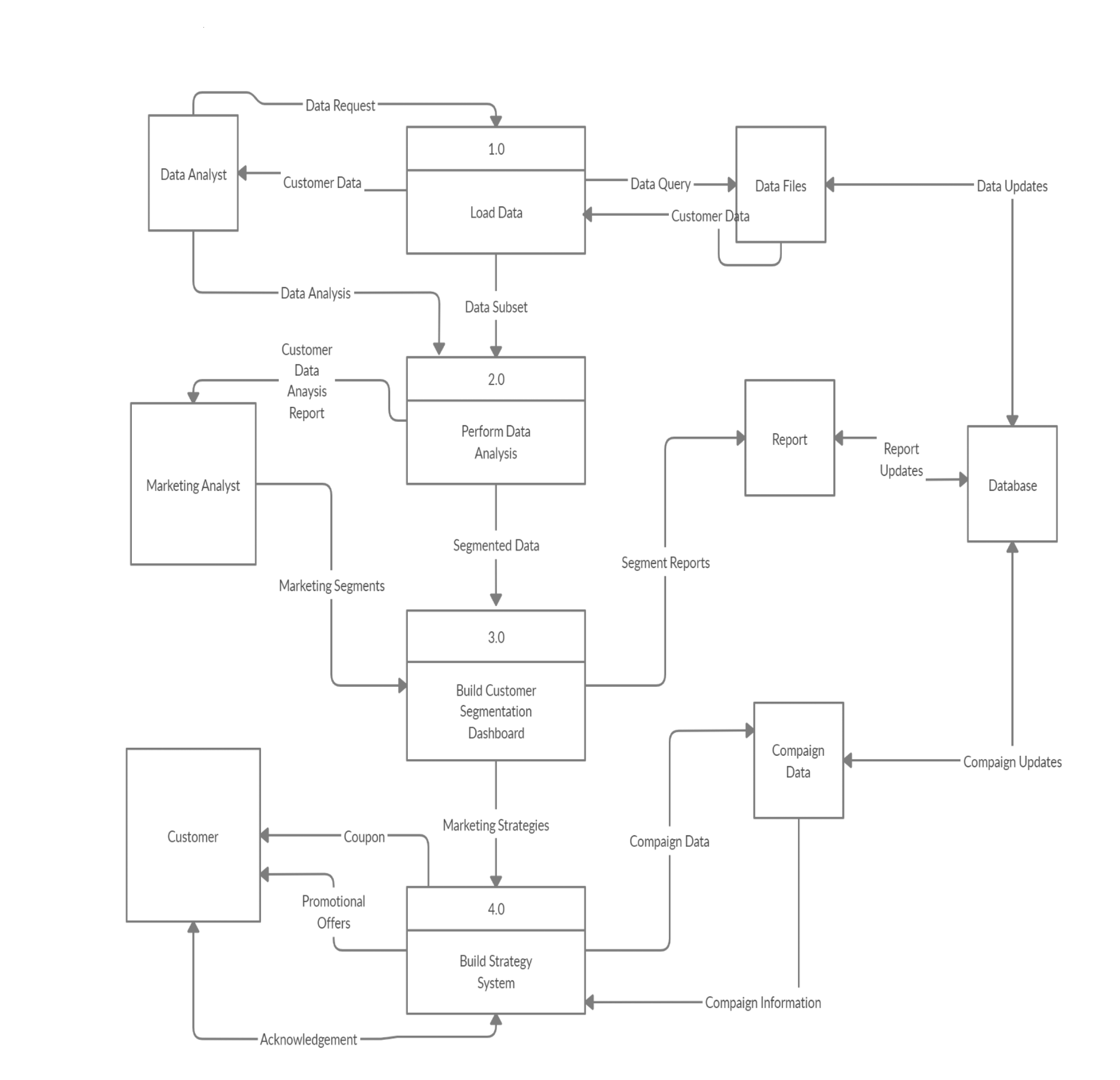}
      \caption{Processes of Customer Segmentation}
      \label{fig:Processes-of-Customer-Segmentation}
\end{figure}

The architecture of the profiling customer shown in figure \ref{fig:Processes-of-Customer-Segmentation}

\subsection{Dataset}
"The data utilized for our research was sourced from the Data Flair repository, which encompasses a cross-border dataset that encompasses several key demographic attributes, including age, education level, ID, annual income, marital status, and presence of children in the household."

\begin{table}[h!]
\begin{center}
    \caption{\footnotesize{Attributes of first datasets}}
\begin{adjustbox}{width=0.5\textwidth}
\begin{tabular}{|c|c|}
\hline
\textbf{Serial No.} & \textbf{Attributes} \\ \hline
1 & ID \\  \hline
2 & Year\_Birth \\  \hline
3 & Education \\  \hline
4 & Marital\_Status \\  \hline
5 & Income \\  \hline
6 & Kidhome \\  \hline
7 & Teenhome \\  \hline

\end{tabular}
\end{adjustbox}
\end{center}
\end{table}

The RFM model employed in this study utilized data from the SAS Institute to calculate the recency, frequency, and monetary rankings, enabling the segmentation of customers into distinct groups. The data comprises the following attributes:

\begin{table}[h!]
\begin{center}
\caption{\footnotesize{Attributes of second datasets}}
\begin{adjustbox}{width=0.5\textwidth}
\begin{tabular}{|c|c|}
\hline
\textbf{Serial No.} & \textbf{Attributes} \\ [0.5ex]
\hline
1 & Dt\_Customer \\ \hline
2 & Recency \\ \hline
3 & MntWines \\ \hline
4 & MntFruits \\ \hline
5 & MntMeatProducts \\ \hline
6 & MntMeatProducts \\ \hline
7 & MntFishProducts \\ \hline
8 & MntSweetProducts \\ \hline
9 & MntGoldProds \\ \hline
10 & NumDealsPurchases \\ \hline
11 & NumWebPurchases \\ \hline
12 & NumCatalogPurchases \\ \hline
13 & NumStorePurchases \\ \hline
14 & NumWebVisitsMonth \\ \hline
15 & AcceptedCmp1 \\ \hline
16 & AcceptedCmp2 \\ \hline
17 & AcceptedCmp3 \\ \hline
18 & AcceptedCmp4 \\ \hline
19 & AcceptedCmp5 \\ \hline
20 & Complain \\ \hline
21 & Z\_CostContact \\ \hline
22 & Z\_Revenue \\ \hline
23 & Response \\ [0.5ex] \hline
\end{tabular}
\end{adjustbox}
\end{center}
\end{table}

\subsection{Preprocessing}
\subsubsection{Data Cleaning}
\begin{figure}[ht!]
      \centering
      \includegraphics[width=0.8\textwidth,height=.60\textwidth]{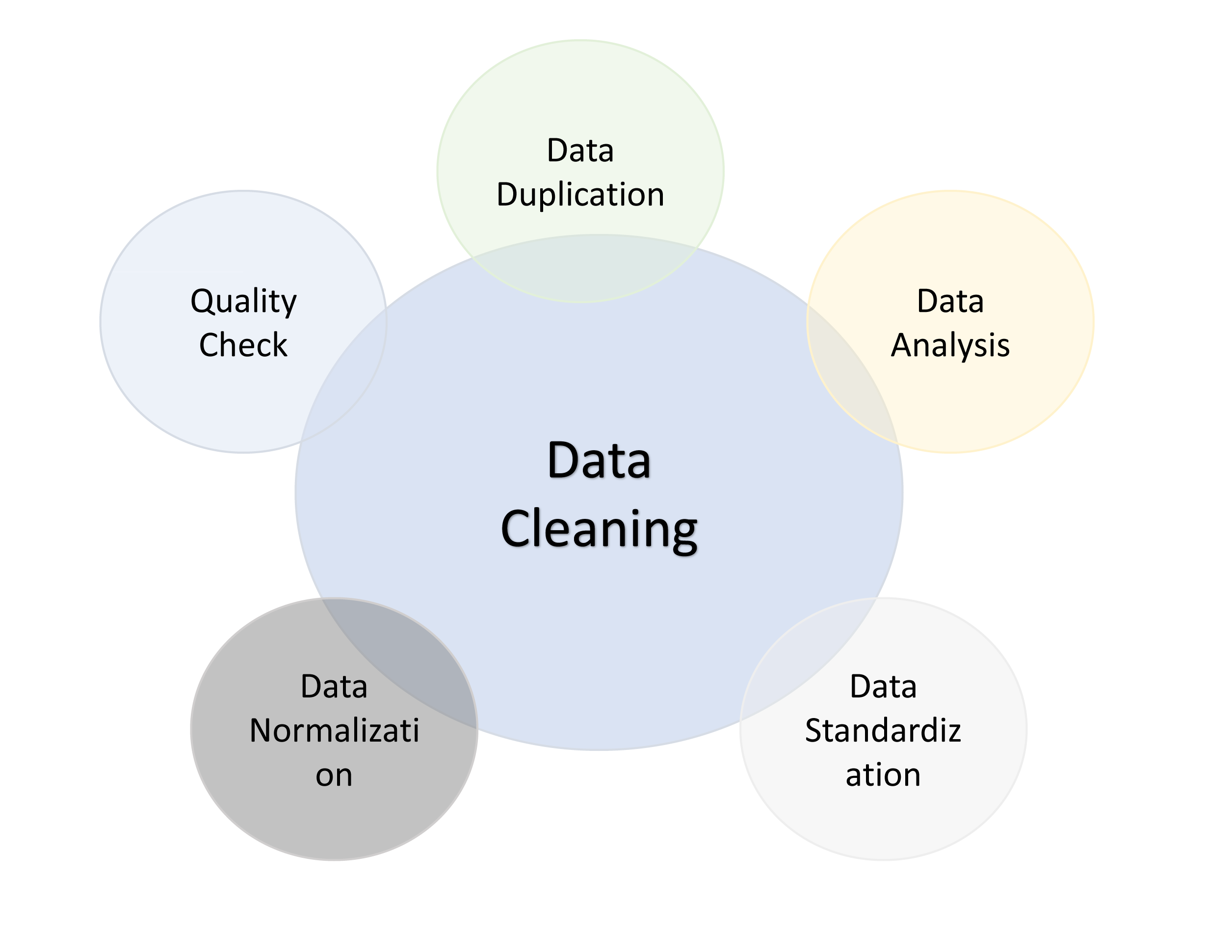}
      \caption{Processes of Customer Segmentation}
      \label{fig:Steps-involved-in-Data-Cleaning}
\end{figure}

Data cleaning is a crucial aspect of AI and has a substantial impact on the development of a model. It is a routine task that is often overlooked, yet it is vital for achieving successful outcomes. While there may not be any complex techniques or insider knowledge involved in data cleaning, it can play a decisive role in the success of a business. Experienced data scientists often allocate a substantial amount of time to this process, recognizing that clean data is more valuable than complex calculations.
With a well-cleaned dataset, even simple calculations can produce desirable results, as demonstrated in Figure \ref{fig:Steps-involved-in-Data-Cleaning}.

\subsubsection{Exploratory Data Analysis}
In the field of data mining, Exploratory Data Analysis (EDA) involves the systematic examination of datasets to uncover their underlying characteristics and patterns. EDA is a crucial step in the data analysis process and helps to understand the information contained in the data before proceeding with modeling. It can be challenging to extract meaningful insights from large sets of raw data or complex calculations. Exploratory data analysis provides a framework for making sense of the data by utilizing visualizations and summarization techniques to make the data more accessible and understandable. The goal of EDA is to provide a comprehensive understanding of the data and identify potential areas for further investigation.

\subsubsection{Analysis of Variables}
Univariate analysis is a fundamental form of data analysis that involves the examination of a single variable. Common univariate techniques include box plots and histograms.
Multivariate analysis, on the other hand, involves the examination of multiple variables. This type of analysis requires the use of more advanced statistical methods such as scatter plots and bar charts.

\subsection{Cluster Analysis}
In this section, the focus of the analysis conducted during this project is presented, providing an overview of segmentation and customer division.

\begin{figure}[ht!]
      \centering
      \includegraphics[width=0.8\textwidth,height=.60\textwidth]{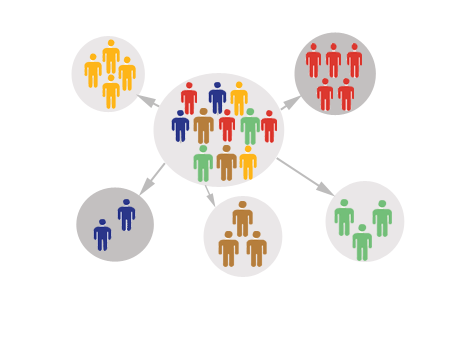}
      \caption{Customer Segmentation}
      \label{fig:Customer-Segmentation}
\end{figure}

Customer division is a widely used marketing strategy that involves dividing the customer base into smaller groups that can be targeted with specific content and offers. These customer segments are drawn from customer behavior data, which provides the business with a deeper understanding of the types of customers in the system. The benefit of customer division is twofold. A better understanding of the types of customers in a system, first and foremost, can lead to better business and marketing strategies. Additionally, a customer is more likely to use an application regularly if they receive relevant content. Furthermore, if a customer is satisfied, they are more likely to recommend the application to others, contributing to the expansion of the business \cite{kashwan2013customer}.
This type of marketing strategy is a component of a company's Business Intelligence framework. To effectively segment the customer base into meaningful groups, a comprehensive analysis of available data along with a study and evaluation of clustering algorithms is necessary (Figure \ref{fig:Customer-Segmentation})

The customer distribution is depicted in Figures \ref{Marital-Status-and-Education-Level}, showcasing the majority of the customers, 64\%, in relationships (Married or Together), with the majority, 97\%, holding at least a bachelor's degree.

\begin{figure}[ht!]
      \begin{subfigure}{0.50\columnwidth}
        \includegraphics[width=\textwidth]{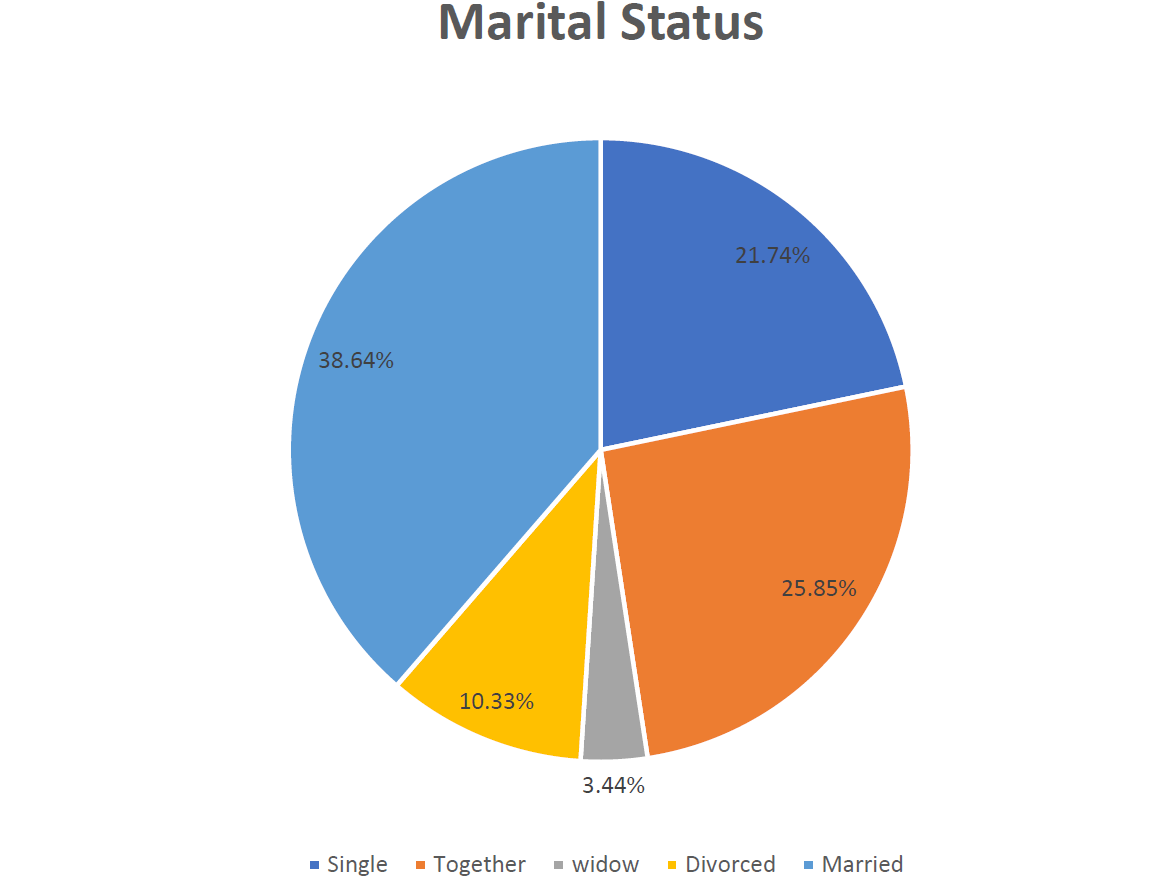}
        \label{fig:Marital-Status}
    \end{subfigure}
    \begin{subfigure}{0.50\columnwidth}
        \includegraphics[width=\textwidth]{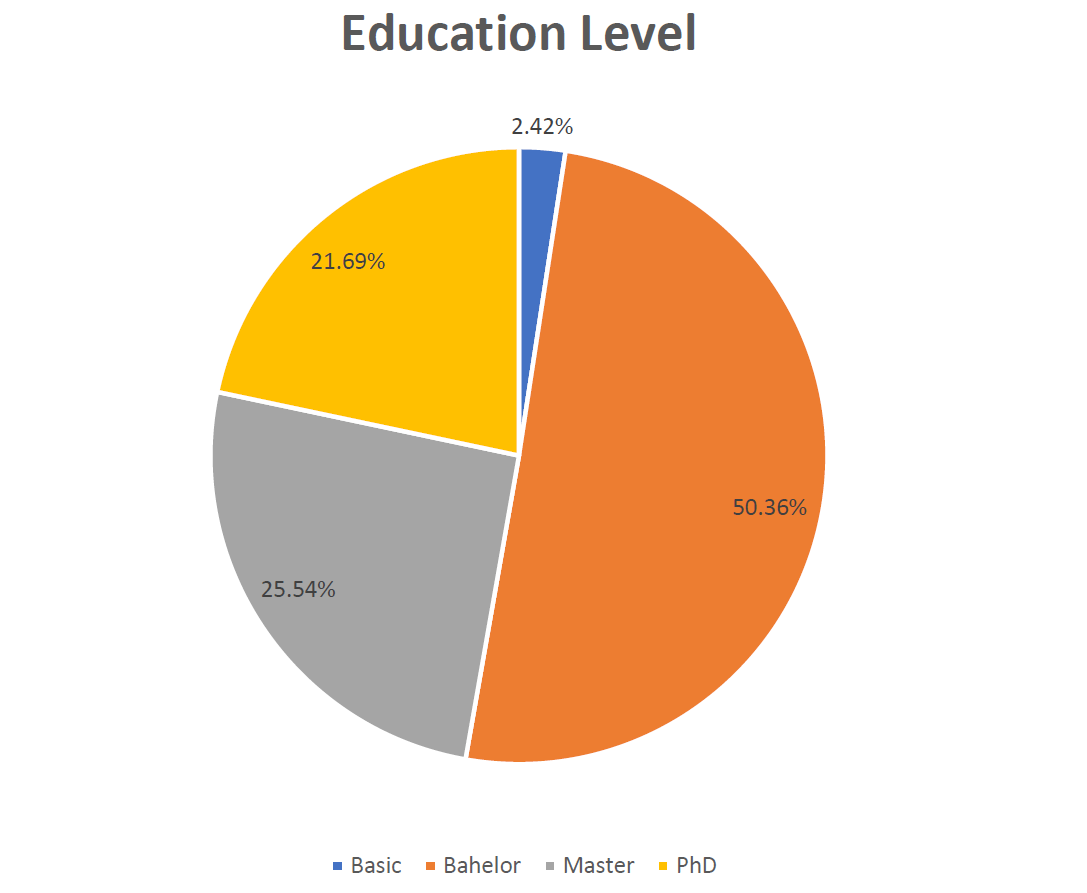}
        \label{fig:Education-Level}
    \end{subfigure}
\caption{Martial Status and Education Level}
\label{fig:Marital-Status-and-Education-Level}
\end{figure}

\subsubsection{Clustering Algorithm}
Clustering algorithms are employed to group clients into clusters in order to ensure that clients belonging to the same cluster are more similar to each other than to clients in another cluster. The aim of this segmentation is to identify meaningful patterns within the data space, with clients being assigned to a cluster based on a chosen distance measure. In this section, we will discuss the most commonly used similarity measures and clustering algorithms. It is important to note that the success of clustering is highly dependent on the definition of a relevant similarity or distance measure.
The simplest and most common distance measure is Euclidian distance (equation 8).

\begin{equation}
\begin{split}
  d(x,y) = \sqrt{\sum_{k=1}^{n} (x_k - y_k)^2 }
\end{split}
\end{equation}

where n is the number of features,
x and y are the data objects,
$x_k$ and $y_k$ are the kth attributes of the feature data objects x and y respectively.

The cosine similarity is widely used in the area of recommender systems, particularly in collaborative filtering. The fundamental concept behind cosine similarity is to calculate the cosine value of the angle between two n-dimensional feature vectors \cite{christy2021rfm}]. This can be accomplished using the following equation, where n represents the number of features in the data objects x and y, denotes the dot product of the vectors, and || x || represents the magnitude of vector x (equation 9).

\begin{equation}
\begin{split}
  cos(x,y) = \frac{(x.y)}{||x| |y||}
\end{split}
\end{equation}

Another distance measure that will be covered in this report is the Pearson correlation. This distance measure is also widely used in recommender systems. Pearson correlation calculates the linear relationship between two feature vectors, meaning two feature vectors are similar if a best-fitting straight line is close to all data points in both vectors. It is calculated using the following function (equation 10):

\begin{equation}
\begin{split}
  Pearson(x,y) = \frac {\sum (x,y)} { \sigma x . \sigma y }
\end{split}
\end{equation}

where x and y are two component vectors,$\sum$ is the covariance of the information focuses x and y, and is the standard deviation of an element vector. The outcome is a worth between - 1 and 1 where a worth near 1 or - 1 implies that all values are situated on the best-fitting line, and values more like 0 show that there is little relationship between the given element vectors.

\subsubsection{K-means}
The K-means algorithm is widely used in cluster analysis and customer segmentation. It is a method designed to divide a set of objects into K subgroups or clusters. The algorithm depends on a pre-determined value for K, with K centroids initialized to random observations within the dataset. The K-means algorithm then iteratively adjusts these centroids to minimize the cluster variance by employing two steps:
\begin{itemize}
    \item For each centroid c, identify the subset of objects that are closer to c than any other centroid using some similarity measure,
    \item Recalculate another centroid for each cluster by computing the mean vector of all objects in the group.
    This two-step process is repeated until convergence is reached. The standard implementation of K-means uses the Euclidean distance measure described in a previous section to identify the subset of objects that corresponds to each cluster, by calculating the mean squared error, which in this case is equivalent to the Euclidean distance, of each object's feature vector to the K centroid and selecting the closest result \cite{kashwan2013customer}. However, other distance measures can be used in place of the Euclidean distance. Aggarwal et al. assert that for high-dimensional data, the choice of distance measure used in clustering is crucial for its success.
    
The steps for the algorithm are as follows:
    \item Choose the number of clusters "k"
    \item Select "k" random points from the dataset as centroids
    \item Assign all points to the nearest group centroid
    \item Recalculate the centroids of newly formed groups
    \item Repeat steps 3 and 4 until no change in clusters is observed
    \item End the iteration.
\end{itemize}

\begin{figure}[ht!]
      \centering
      \includegraphics[width=0.5\textwidth,height=.60\textwidth]{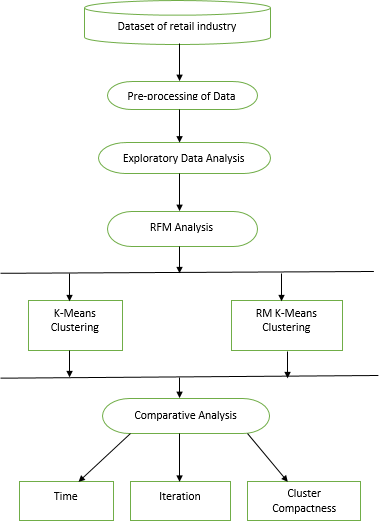}
      \caption{Processes of RFM analysis}
      \label{fig:Processes-of-RFM-analysis}
\end{figure}

Process of K-Means on RFM analysis shown \ref{fig:Processes-of-RFM-analysis}).

.

\subsubsection{Silhouette Score}
This is a more accurate way of determining the clustering results to form from the data. It is determined for each case, and the formula is as follows (11):

\begin{equation}
\begin{split}
  SC = \frac {x-y} { max(x,y)}
\end{split}
\end{equation}
where y represents the mean cluster formation distance, or the distance between examples in the same cluster, and x represents the mean nearest cluster distance, or the distance between instances in the next closest cluster.

\begin{figure}[ht!]
      \centering
      \includegraphics[width=0.6\textwidth,height=.40\textwidth]{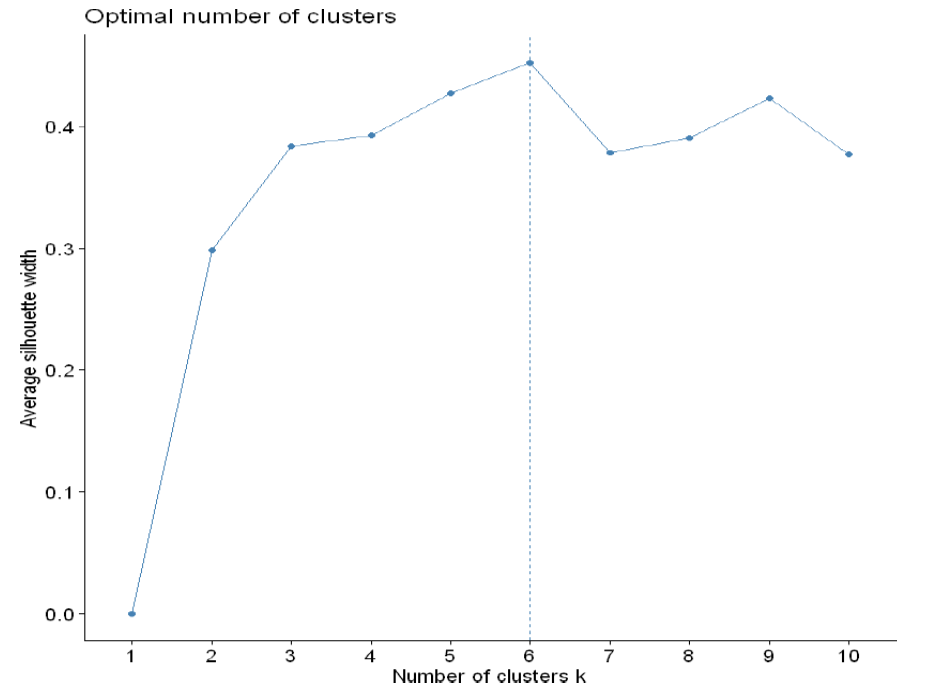}
      \caption{Silhouette Method}
      \label{fig:Silhouette-Method}
\end{figure}

The coefficient ranges from -1 to 1. A number around 1 indicates that the example is near its cluster and belongs to the correct cluster. A high ratio to -1, on the other hand, indicates that the value has been assigned to the incorrect cluster (Figure \ref{fig:Silhouette-Method}).

This approach is predicated on the assumption that k=3 is a local optimum, while k=5 should be selected as the number of clusters. This method is deemed to be superior as it renders the determination of the optimal number of clusters more critical and transparent. However, it should be noted that this calculation is computationally intensive, as the coefficient must be computed for each case \cite{rousseeuw1987silhouettes}. As such, the choice regarding the optimal metric for selecting the number of clusters must be made based on the specific requirements of the application.

\subsubsection{Elbow method}
\begin{figure}[ht!]
      \centering
      \includegraphics[width=0.6\textwidth,height=.40\textwidth]{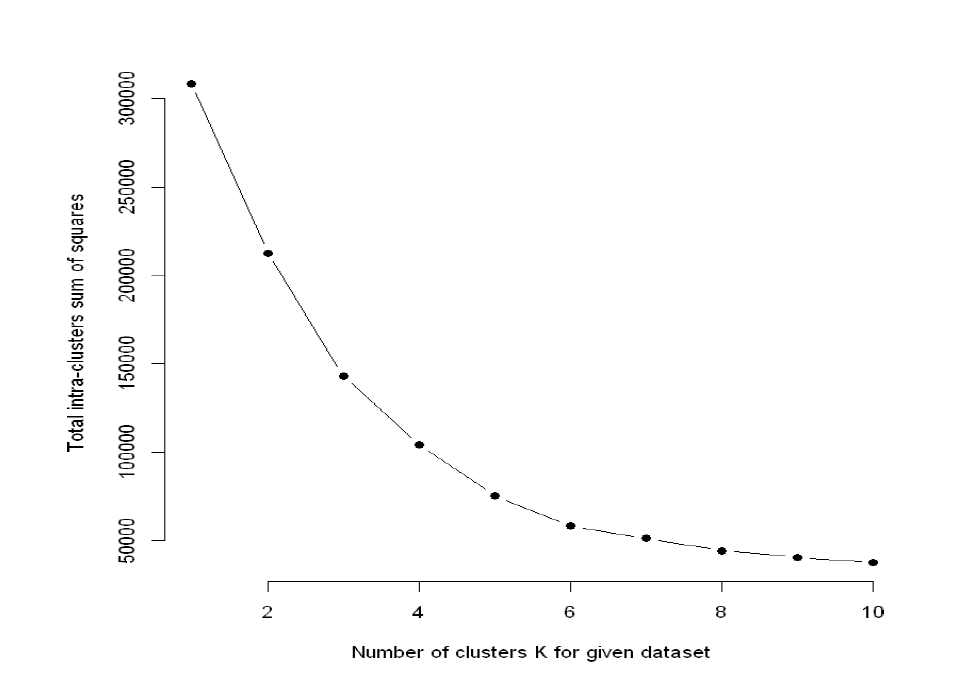}
      \caption{Elbow Method}
      \label{fig:Elbow Method}
\end{figure}

\begin{figure}[ht!]
      \centering
      \includegraphics[width=0.6\textwidth,height=.40\textwidth]{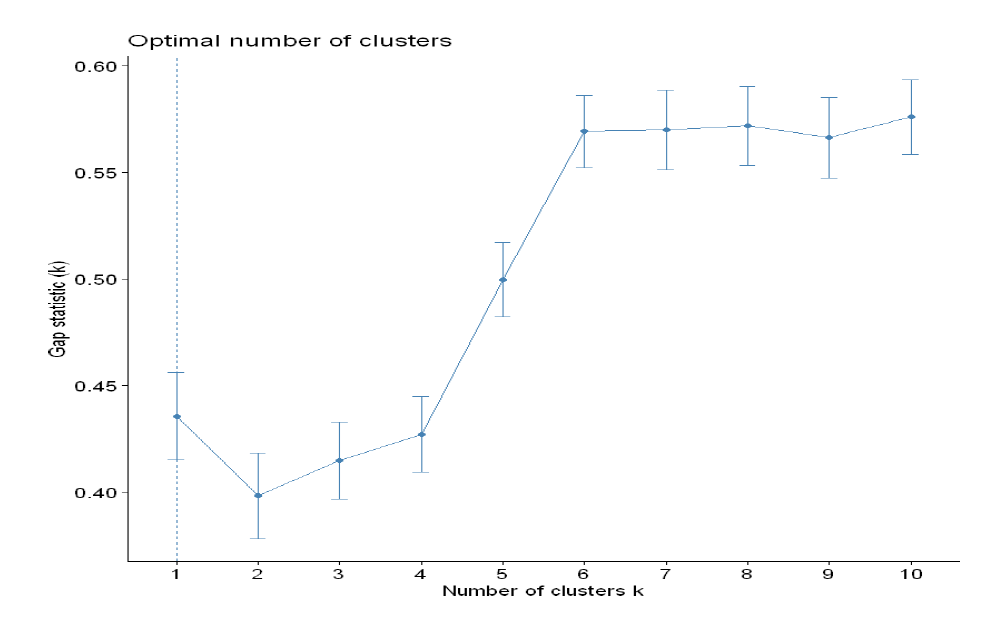}
      \caption{Gap Statistics Method}
      \label{fig:Gap Statistics Method}
\end{figure}

The basic purpose of cluster partitioning algorithms such as k-means is to define clusters with the least amount of intra-cluster variation (Figure \ref{fig:Elbow Method}). Minimum( summary W($C_k$), k=1 to k shown (Figure \ref{fig:Gap Statistics Method}).

where W($C_k$) denotes intra-cluster variation, $C_k$ denotes the kth cluster. The compact of the clustering boundary can be assessed by measuring the total intra-cluster variation.

In 2001, the Gap Statistic Method was introduced by G.Walther, R. Tibshirani, and T. Hastie of Stanford University. This method can be applied to any clustering technique, such as K-means, hierarchical clustering, etc. The gap statistic allows us to evaluate the total intra-cluster variation for different values of k, along with their average values, under the assumption of an uninformative reference distribution of the data. Monte Carlo simulations can be used to generate a sample dataset. For each variable in the dataset, we can calculate the range between the minimum and maximum values, from which we can generate values uniformly distributed within the lower and upper bounds.

The information investigation stage, this is a lopsided dataset (more than 80\% express no to the mission). The models are not difficult to become familiar with certain characteristics about regrettable examples, yet it very well may be difficult to get from positive examples.
While SMOTE lighten the issue by offering us more certain preparation tests.
Simultaneously,Matthews Correlation Coefficient (MCC) scorer considers valid and misleading up-sides and negatives and is for the most part viewed as a reasonable measure which can be utilized regardless of whether the classes are of totally different sizes. For this situation, MCC is a more effective measure than exactness in trial, since there are a couple of positive examples in the test set.
we have tried Logistic Regression, The Boosting Tree, Support-vector machines ,and Neural Networks.
Supporting Tree plays out the best among every one of the models in each of the 3 datasets. The exhibitions of SVM and NN are practically equivalent in 3 different datasets. LR is the most exceedingly terrible model since it is excessively basic for this grouping task.
Supporting Tree have a few anomalies in crude datasets and include choice datasets, which demonstrates this calculation probably won't be steady in these datasets.
All in all, in this characterization task, we could utilize Feature Selection Dataset + Boosting Tree, in light of the fact that
1) This mix accomplishes the best MCC execution.
2) Although BT may be unsteady, even the lower exceptions are practically identical to NN and SVM.

The overall accuracy of the model was determined to be 0.877, but upon further examination of the score report, it was noted that the model performed well in identifying negative examples (0) with high accuracy (0.877), however it was found to be lacking in its ability to accurately identify certain positive examples, with an accuracy of 0.55 and recall of 0.55.

The MCC for the test data was calculated to be 0.469, indicating that the model may have difficulty in correctly classifying positive examples in the test set. While the train MCC was determined to be 0.98, this result suggests the presence of overfitting in the model. Despite efforts to improve the model, a reduction in train MCC did not result in significant improvement in test MCC, which may indicate that the features used in this dataset may not effectively predict the 'Response' variable.

\section{CONCLUSION}
Our research paper investigates the formation of a client profile and the prediction of the client's behavior. For a precise forecast, regression analysis requires different client characteristics and the time series needs to consider the client's purchase history. To determine the market area and create a client profile, segmentation types and client-characterizing variables are utilized.

Response modeling is commonly framed as a binary classification problem. Buyers are divided into two categories: responders and non-responders. Various classification techniques, such as statistical methods and AI methods, were employed to model the response, including decision trees, Bayesian networks, and support vector machines. The latter of these, support vector machine (SVM), has gained attention in the AI community and offers advantages over multivariate classifiers. In this study, a support vector machine (SVM) was employed as a classifier for the simulation.

The response modeling process involves several steps, including data acquisition, data preprocessing, feature engineering, feature selection, class balancing, model training and evaluation. Different data mining techniques and algorithms were employed to execute each step. This study utilized the analysis cycle, which was built upon previous response modeling philosophy. Due to the nature of this study, different stages of the modeling process were gathered from past work and, with certain modifications and enhancements, integrated into a single process. To choose the best algorithm and philosophy for each stage of the cycle, various studies related to each stage were considered and evaluated. After considering various techniques and approaches for each stage, the best and most appropriate ones were selected. This exploratory process (building the response model) required extensive programming to implement. The algorithms and methods associated with each stage were customized and run using the Python programming language.

In considering the limitations of the project, it is important to take into account the amount of data available, as well as potential avenues for future research and the possibility of extending the project. The amount of data, specifically the number of rows in this study, has a direct impact on the accuracy and solution provided by the algorithm. With a limited dataset of under 3000 rows, the results may not be fully representative. Additionally, this study did not consider interpretability, which is an area of interest in the field of customer segmentation based on distinct categories. Companies would benefit from being able to understand why their clients are purchasing certain goods rather than simply predicting future purchases. While it may be possible to extract the importance of certain variables from some of the algorithms used in this study, it was not within the scope of this project's objective.

\section{Future Work}
In future Work, more advanced methods for predicting customer churn may be explored, such as weighted random forests and hybrid models that can handle unstructured data. This would enable the extraction of relevant attributes for potential customer segmentation studies in the retail industry. As highlighted in the literature review, using hybrid models has shown promising performance gains and could be a strategy to improve the models.

Artificial intelligence has the potential to revolutionize various industries by transforming existing business processes and creating new business models. Key areas of focus include consumer engagement, digital manufacturing, smart cities, autonomous vehicles, risk management, computer vision, and speech recognition. AI has already demonstrated positive results in a range of sectors including healthcare, law enforcement, finance, security, trade, manufacturing, education, mining, and logistics.

 \bibliographystyle{elsarticle-num} 
 \bibliography{elsarticle-template-num}





\end{document}